\documentclass[10pt,twocolumn,letterpaper]{article}

\usepackage{cvpr}              
\usepackage[dvipsnames]{xcolor}

\definecolor{cvprblue}{rgb}{0.21,0.49,0.74}
\usepackage[pagebackref,breaklinks,colorlinks,citecolor=cvprblue]{hyperref}
\usepackage{multirow}
\def\paperID{368} 
\def\confName{3DV\xspace}
\def\confYear{2024\xspace}

\title{Out of the Room: Generalizing Event-Based Dynamic Motion Segmentation for Complex Scenes}

\author{ Stamatios Georgoulis$^1$\footnotemark[1] \qquad Weining Ren$^2$\footnotemark[1]  \qquad Alfredo Bochicchio$^1$\qquad  Daniel Eckert$^1$\\ 
\qquad Yuanyou Li$^1$ \qquad Abel Gawel$^1$\footnotemark[2]\\ 
$^1$Huawei Technologies, Zurich Research Center \qquad $^2$ETH Zurich \\ 
{{\tt \small $^1$ stamatios.georgoulis@huawei.com}\qquad {\tt \small$^2$ weiren@ethz.ch}}}

\begin{document}
\twocolumn[{%
\renewcommand\twocolumn[1][]{#1}%
\maketitle
\vspace{-3em}
\begin{center}
    \centering
    \captionsetup{type=figure}
    \includegraphics[width=\textwidth,height=5cm]{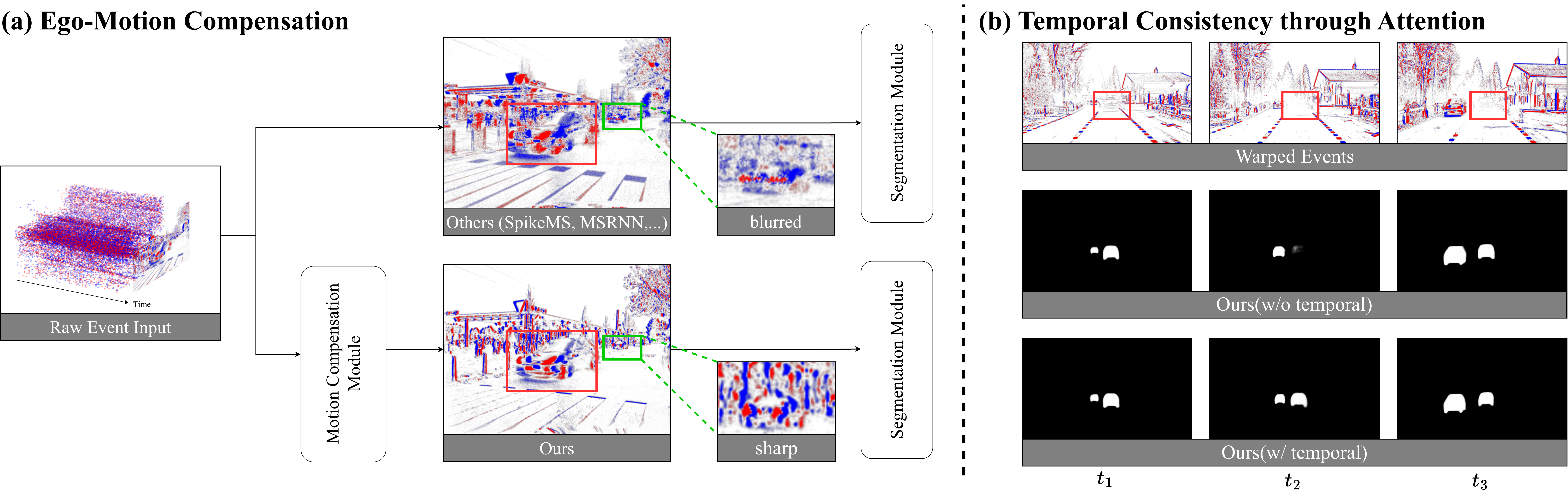}
    \captionof{figure}{In this paper, we generalize event-based motion segmentation to large-scale outdoor scenes. To realize this, we propose two key features: (a) Motion Compensation. Unlike current works \cite{zhang2023multi, mitrokhin2019ev} that utilize the raw event representation (e.g. voxel grid) as input to motion segmentation and let the network figure out both ego-motion and dynamic object motion at once, we argue that ego-motion compensating the event representation (by predicting depth and 6DoF pose) is a necessary pre-processing step to motion segmentation, as it makes static regions sharper (green box) while leaving dynamic regions blurry (red box); b) Temporal Attention. Due to the inherent noisy and jittery nature of events, which can disappear and re-appear between adjacent time steps (red box), it is crucial to incorporate temporal consistency modules into motion segmentation. Together, these features greatly boost the system's overall performance.}
    \label{fig:teaser}
\end{center}%
}]

\begin{abstract}
\def\thefootnote{*}\footnotetext{Equal contribution.}
\def\thefootnote{†}\footnotetext{Corresponding Author.}
Rapid and reliable identification of dynamic scene parts, also known as motion segmentation, is a key challenge for mobile sensors. Contemporary RGB camera-based methods rely on modeling camera and scene properties however, are often under-constrained and fall short in unknown categories. Event cameras have the potential to overcome these limitations, but corresponding methods have only been demonstrated in smaller-scale indoor environments with simplified dynamic objects. This work presents an event-based method for class-agnostic motion segmentation that can successfully be deployed across complex large-scale outdoor environments too. To this end, we introduce a novel divide-and-conquer pipeline that combines: (a) ego-motion compensated events, computed via a scene understanding module that predicts monocular depth and camera pose as auxiliary tasks, and (b) optical flow from a dedicated optical flow module. These intermediate representations are then fed into a segmentation module that predicts motion segmentation masks. A novel transformer-based temporal attention module in the segmentation module builds correlations across adjacent 'frames' to get temporally consistent segmentation masks. Our method sets the new state-of-the-art on the classic EV-IMO benchmark (indoors), where we achieve improvements of 2.19 moving object IoU (2.22 mIoU) and 4.52 point IoU respectively, as well as on a newly-generated motion segmentation and tracking benchmark (outdoors) based on the DSEC event dataset, termed DSEC-MOTS, where we show improvement of 12.91 moving object IoU.
\end{abstract}

\section{Introduction} 

Fast dynamic environments present significant challenges for mobile perception. Under such conditions, it is crucial for mobile devices to differentiate between dynamic and static scene contents, known as motion segmentation, for a wide range of applications. Dynamic objects corrupt pose estimation algorithms and scene reconstruction \cite{cadena2016past, rosinol2021kimera}, disturb ego-motion estimation and localization from exteroceptive sensors \cite{akai20203d}, or challenge navigation \cite{eppenberger2020leveraging}. 

In the literature, the challenge of motion segmentation has been mainly studied from the perspective of a moving RGB camera \cite{xie2022segmenting, bao2022discovering, yang2021learning, bideau2018moa, bideau2016s}, and particularly as a side-task for improving structure-from-motion pipelines in dynamic scenes \cite{li2021unsupervised, bian2021unsupervised, klingner2020self, casser2019unsupervised, liu2019unsupervised}. These approaches typically train one or more neural networks (often complemented by optimization techniques) that implicitly or explicitly model 2D (e.g. optical flow, depth, instance segmentation) or 3D (e.g. pose, scene flow) properties of the camera and scene, before reasoning about what is dynamic. Understandably, this is a cumbersome and frequently under-constrained task in the RGB space \cite{yang2021learning} that presumes full scene understanding. 


Differently from RGB-based approaches, another line of works have employed event cameras to tackle motion segmentation. Event cameras \cite{gallego2020event}, like DVS \cite{lichtsteiner2008128} or DAVIS \cite{brandli2014240}, are bio-inspired sensors that asynchronously capture per-pixel, positive or negative, log-brightness changes in the form of events $\epsilon_{i}=\{x_{i},y_{i},t_{i},p_{i}\}$, with $x_{i},y_{i}$ denoting the spatial coordinates, $t_{i}$ the timestamp, and $p_{i}$ the polarity of the triggered event. Due to their working principle event cameras can capture motion with dense temporal resolution (in the order of microseconds), in high dynamic range, and without blur. Naturally, these unique characteristics render event cameras a good fit for the motion segmentation problem. However, existing event-based motion segmentation works, either optimization-based \cite{parameshwara20210, he2021fast, zhou2021event, falanga2020dynamic, stoffregen2019event, mitrokhin2018event} or learning-based \cite{zhang2023multi, parameshwara2021spikems, mitrokhin2020learning, mitrokhin2019ev}, do not fully leverage the strengths of event cameras, e.g. handle arbitrary forms of ego- and scene motion for motion segmentation and have been shown to work mainly on indoor datasets with simplified examples as dynamic objects \cite{mitrokhin2020learning}.


In this paper, we introduce a learning-based motion segmentation approach solely from events that works on complex dynamic scenes with arbitrary moving objects, as well as joint camera and object motions. Our starting point is the key observation that current learning-based works \cite{zhang2023multi, parameshwara2021spikems, mitrokhin2020learning, mitrokhin2019ev} use events in their raw format, and let the neural network(s) disambiguate between events caused by: (a) camera ego-motion, or (b) dynamic objects. We show that this is a sub-optimal design, and instead propose to motion compensate the raw events \cite{gallego2018unifying}, i.e. remove their ego-motion component, before using them for motion segmentation. Figure \ref{fig:teaser}(a) illustrates why this is a good practice: We observe that after ego-motion compensation, events caused by (a) become sharp, while the ones caused by (b) remain blurry. Still, in challenging scenarios, like far away objects with relatively small motions, even ego-motion compensated events are not enough for accurate motion segmentation. Hence, we propose to augment the ego-motion compensated events with dense optical flow computed from the raw events to help the neural network better reason in such scenarios. Finally, we notice that the predicted segmentation masks tend to jitter over time since events do not trigger consistently across `frames' in real-life scenarios, as shown in Figure \ref{fig:teaser}(b). In order to rectify temporally inconsistent predictions, we propose to incorporate into our motion segmentation pipeline temporal information from previous time steps in the form of hidden neural states.

This work pioneers event-based motion compensation in unconstrained environments with these main contributions: (a) It uses a divide and conquer approach that first ego-motion compensates the raw events, which, together with optical flow, are fed into (b) a novel motion segmentation model that employs temporal attention for temporally consistent motion segmentation masks, (c) establishing a new state-of-the-art on the public EV-IMO benchmark by a large margin (+4.52 pIoU) and the newly-generated DSEC-MOTS dataset (+12.91 IoU). 

\section{Related Work}

Although the problem of motion segmentation has also been studied from the perspective of a monocular RGB camera, e.g. \cite{xie2022segmenting, bao2022discovering, yang2021learning, bideau2018moa, bideau2016s}, in this section, we focus on works that instead use a monocular \textit{event camera} \cite{gallego2020event} to tackle this task. In the following paragraphs, we first discuss event-based approaches that are most related to our work, i.e. motion segmentation, and gradually move to loosely related tasks, i.e. scene understanding.

\paragraph{Motion Segmentation} In general, motion segmentation from event cameras can be divided into two groups. First, classic \textit{optimization-based} methods \cite{parameshwara20210, he2021fast, zhou2021event, falanga2020dynamic, stoffregen2019event, mitrokhin2018event} where the desired hyper-parameters, typically the event clusters, the motion of each event cluster, the number of event clusters, etc., are optimized on a per-sample basis to minimize the employed loss function(s). Second, \textit{learning-based} methods \cite{zhang2023multi, parameshwara2021spikems, mitrokhin2020learning, mitrokhin2019ev} that train the weights of the designed model, typically a Convolutional Neural Network (CNN), a Graph Neural Network (GNN), a Spiking Neural Networks (SNN), etc., on large-scale datasets and then deploy the trained model on new samples during inference.  

Regarding optimization-based methods, \citet{mitrokhin2018event} used an event representation based on average timestamps to fit a parametric model that describes the camera motion, and detected as moving objects events that do not conform to that model. \citet{falanga2020dynamic} adapted this concept for quadrotors, but opted for runtime efficiency by estimating rotational camera motion from IMU, and detected moving objects by thresholding the average timestamps event representation. \citet{stoffregen2019event} leveraged the contrast maximization objective function \cite{gallego2018unifying} and proposed to alternatively optimize the event clusters and their motion parameters to separate events of individually moving objects or the background. Similarly, \citet{zhou2021event} optimized for event clusters and their motion parameters, but differently, they relied on graph cuts to incorporate spatio-temporal constraints on the events. \citet{he2021fast} combined an event camera, a depth camera, and IMU to ego-motion compensate events, construct an average timestamps event representation, and use it to detect, localize dynamic objects, and track them over time. Finally, \citet{parameshwara20210} introduced a pipeline that combines feature tracking and motion compensation, to segment a dynamic scene into motion clusters following a `splitting and merging' algorithm, as well as a `motion propagation and cluster keyslices' algorithm for further speed-up.

In general, the aforementioned optimization-based methods have certain advantages: (a) they do not require large-scale datasets with expensive ground truth annotations for training; (b) they can learn class-agnostic motion clusters. At the same time, however, these methods come with several disadvantages: (i) they are relatively slow algorithms (e.g. 4s in \cite{zhou2021event}), especially for safety-critical applications like autonomous driving; (ii) they make strong assumptions, like the number of motion clusters, the type of motion (i.e. mostly rotational motions, as translational ones require depth for motion compensation), etc.; (iii) they require cumbersome hyper-parameter tuning for optimal results, which usually comes together with low generalization capability outside the tuned hyper-parameter space; (iv) they rely on delicate optimization procedures that are prone to degenerate solutions or local minima. 

Regarding learning-based methods, \citet{mitrokhin2019ev} proposed an adaptation of the structure-from-motion CNN \cite{zhou2017unsupervised} to learn depth, 6DoF camera pose, per-pixel segmentation masks, and per-object 3D translational velocities. They also introduced the EV-IMO dataset for event-based motion segmentation, featuring indoor scenes with (mostly rotational) camera motion and up to 3 moving objects. Most recently, \citet{zhang2023multi} proposed a multi-scale recurrent CNN to leverage long-range temporal information when learning the motion masks. Beyond CNNs, \citet{mitrokhin2020learning} employed a GNN and \citet{parameshwara2021spikems} an SNN to tackle the problem of motion segmentation. 

Typical advantages of the described learning-based methods are: (a) they have faster runtimes, that are restricted only by the network design; (b) they show robustness with respect to the number of dynamic objects, and the type of camera motion (rotational and translational); (c) they may come with stronger generalization capabilities in out-of-distribution samples. Their disadvantages can be summarized as: (i) They have been shown to work only on indoor scenes with a small number of toy dynamic objects (EV-IMO)\footnote{Only \cite{zhou2023dsec} is known to work on complex outdoor scenes (DSEC-MOD), but it focuses on detection of dynamic objects, rather than segmentation as we do, and furthermore it uses information from both RGB and events cameras as input.}; (ii) except \cite{zhang2023multi}, they do not take the temporal consistency of segmentation masks into account, leading to inconsistent predictions among neighboring `frames'. Our approach belongs to this group of works too, but it comes with key algorithmic novelties that lead to significant performance boost, i.e. (1) we propose a divide-and-conquer approach where we first motion compensate the raw events \cite{gallego2018unifying} before using them for motion segmentation; (2) to rectify temporally inconsistent predictions, we incorporate temporal information from previous time steps in the form of hidden neural states, but unlike \cite{zhang2023multi} that fuses them using LSTM modules, we propose a transformer-based temporal attention module. These novelties enable our method to work on complex outdoor scenes with arbitrary dynamic objects and motion types (DSEC-MOTS), while still achieving significant improvements on simpler indoor scenes with few toy dynamic objects (EV-IMO).

\paragraph{Scene Understanding} From a general point of view our method incorporates scene understanding tasks on its route to handle motion segmentation, i.e. monocular depth, 6DoF pose and optical flow estimation. Although in the event-based literature several works tackle these problems individually, e.g. monocular depth \cite{hidalgo2020learning, ye2020unsupervised, zhu2019unsupervised}, optical flow \cite{shiba2022secrets, gehrig2021raft, hagenaars2021self, lee2020spike, pan2020single, zhu2018ev}, pose \cite{liu2021spatiotemporal, nunes2020entropy, liu2020globally, peng2020globally, ye2020unsupervised, zhu2019unsupervised, gallego2018unifying}, semantic segmentation \cite{sun2022ess, wang2021evdistill}, object detection \cite{schaefer2022aegnn, li2021graph, perot2020learning}, the scope of this work is beyond supremacy in each individual sub-task. Instead, we rather use them as a means to achieve better motion segmentation.

\section{Methodology}

\begin{figure*}
  \includegraphics[width=\textwidth]{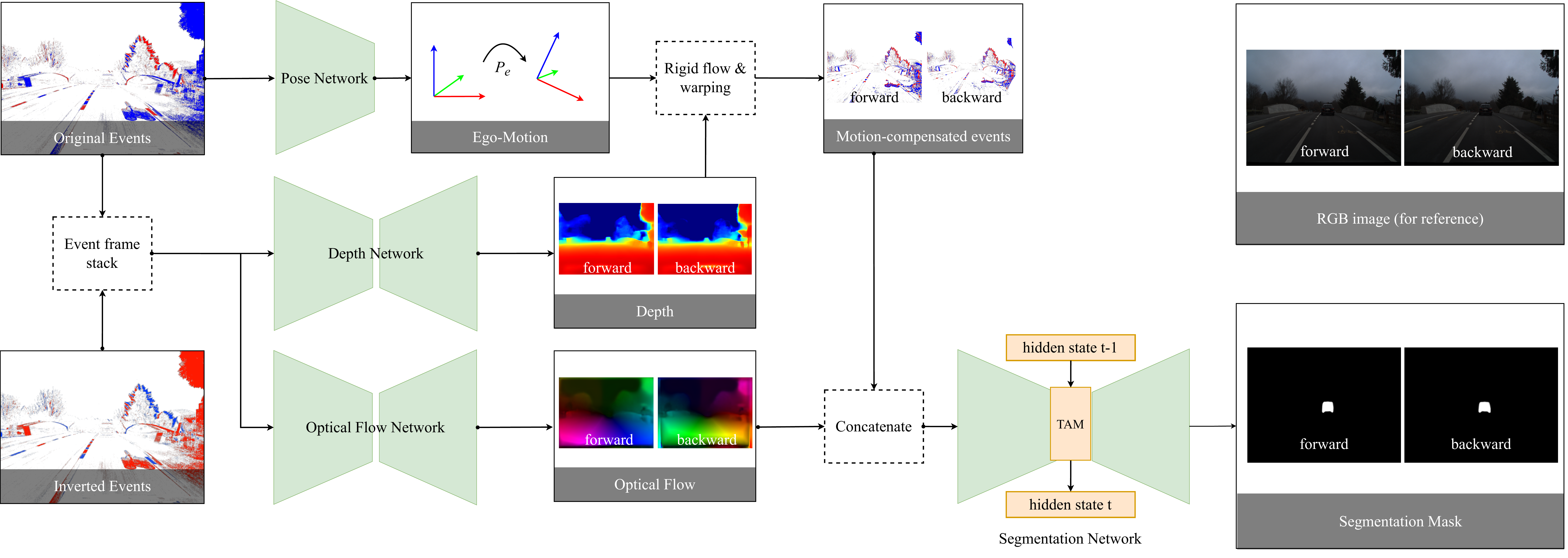}
  \caption{\textbf{System overview}: Our pipeline adopts a divide-and-conquer approach that operates on three steps, namely ego-motion compensation (see Section \ref{subsec:mot_comp}), optical flow estimation (see Section \ref{subsec:flow}, and motion segmentation (see Section \ref{subsec:seg}). First, ego-motion compensated events (backward and forward) are computed by warping the input event representation using the predicted depth maps and 6DoF camera pose. Second, optical flow (backward and forward) is estimated from the input event representation in parallel. Third, the ego-motion compensated events are concatenated with the optical flow and fed as input to the motion segmentation network that predicts the motion segmentation masks (backward and forward). A Temporal Attention Module that applies channel and spatial attention across the hidden states of different timestamps ($t$, $t-1$) is employed inside the motion segmentation network for temporally consistent motion masks.
  }
  \label{fig:pipeline}
\end{figure*}

\subsection{System Overview}

The input to our pipeline is the raw asynchronous event stream $\mathcal{E} = \{\epsilon_{i}\}_{i=1}^{N}$ recorded by a moving event camera in a dynamic scene within a time window $\Delta t$ ($100$ms in our case). In certain learning pipelines, like GNNs \cite{schaefer2022aegnn, mitrokhin2020learning} or SNNs \cite{parameshwara2021spikems}, the raw asynchronous event streams can be used as are. However, in our pipeline we opt to convert the asynchronous event stream into a synchronous representation suitable for training CNNs \cite{gehrig2019end}. In particular, we adopt the Event Frame Stack representation, which splits the time window $\Delta t$ into $B$ equally sized temporal bins ($10$ in our case) and for each bin constructs an $H\times W$ Event Frame \cite{gehrig2019end}, with $H$, $W$ being the spatial dimensions. Finally, all Event Frames are stacked together forming an $B\times H\times W$ Event Frame Stack $S_{e}$ that serves as input to our networks. The output is a binary segmentation mask $M$ at the end of the time window $\Delta t$ that clusters each spatial location into static or moving parts.

Our overall pipeline is depicted in Figure \ref{fig:pipeline}. It consists of three steps, i.e. ego-motion compensation, optical flow estimation, and motion segmentation. For simplicity, below we describe the procedure to predict segmentation masks $M$ at $t_{1}$ (forward), but during training we also predict them at $t_{0}$ (backward)\footnote{In practice, this means estimating depth $D_{e}$ at $t_{0}$, 6DoF pose $P_{e}$ from $t_{0}\rightarrow t_{1}$, and optical flow $F_{e}$ from $t_{0}\rightarrow t_{1}$ by simply reversing the raw asynchronous event stream \cite{tulyakov2021time} as shown in Figure \ref{fig:pipeline}}. Within the time window $\Delta t = t_{1} - t_{0}$, our \textit{ego-motion compensation} step estimates monocular depth $D_{e}$ (at $t_{1}$) and 6DoF pose $P_{e}$ (from $t_{1}\rightarrow t_{0}$) from the event representation $S_{e}$, which are used to ego-motion compensate the event representation $S_{e}^{mc}$ (at $t_{1}$). That is, geometrically warp each Event Frame into a single timestamp ($t_{1}$ in our case). In parallel, the \textit{optical flow} step is producing optical flow $F_{e}$ (from $t_{1}\rightarrow t_{0}$) using the event representation $S_{e}$. Finally, ego-motion compensated events $S_{e}^{mc}$ (at $t_{1}$) and optical flow $F_{e}$ (from $t_{1}\rightarrow t_{0}$) are concatenated and fed into our \textit{motion segmentation} step that produces the final segmentation mask $M$ (at $t_{1}$). Note that, our motion segmentation network in its bottleneck layer also incorporates temporal information from previous timestamps as hidden states. Below, we describe each step in more detail.

\subsection{Ego-Motion Compensation} \label{subsec:mot_comp}

As mentioned before, a key algorithmic novelty of our pipeline is that we opt for ego-motion compensating the event representation $S_{e}$, before using it for motion segmentation\footnote{Bypassing this step, as in \cite{zhang2023multi, mitrokhin2020learning}, essentially forces the motion segmentation network to concurrently solve two tasks: (1) reason about the ego-motion component, and (2) classify which events are caused by it, and which by dynamic objects; this is arguably a tougher problem.}. In the case of rotational-only camera motion \cite{zhou2021event, falanga2020dynamic, stoffregen2019event, mitrokhin2018event}, ego-motion compensation can be achieved by estimating the 3DoF pose change $P_{e}$ (a rotation matrix here) for the whole time window $\Delta t$, and based on it warp the event representation $S_{e}$ at the temporal bins level. Assuming that $p_{mc}$ denotes the homogeneous spatial coordinates of a pixel in the ego-motion compensated event representation $S_{e}^{mc}$, $K$ is the event camera intrinsic matrix, and $p(b)$ indicates the homogeneous spatial coordinates of a pixel in the event representation $S_{e}$ for temporal bin $b$ out of $B$, then the relation between $p(b)$ and $p_{mc}$ is given by the following equation:
\begin{equation} \label{eq:1}
    p(b) \sim K \cdot P_{e}(b) \cdot K^{-1} \cdot p_{mc},
\end{equation}
with $P_{e}(b)$ being a linear interpolation of the total 3DoF pose change $P_{e}$ at temporal bin $b$. Note that, the projected spatial coordinates $p(b)$ contain continuous values, hence we need to apply differentiable bi-linear sampling \cite{jaderberg2015spatial} in order to backward warp $p(b)$ to $p_{mc}$. Also, note that $p_{mc}$ receives multiple $p(b)$ values, once for every temporal bin $b$, which are consequently summed up to form the ego-motion compensated event representation $S_{e}^{mc}$ at the desired timestamp ($t_{1}$ here). 

In this paper, we are interested in the more general case where both rotational and translational camera motion are present \cite{gehrig2021dsec}, and as a result, ego-motion compensation requires to estimate the 6DoF pose change $P_{e}$ (a rotation matrix and a translation vector here), but also the depth values $D_{e}$ at the timestamp where ego-motion compensation is performed ($t_{1}$ here). Given the latter, Equation \ref{eq:1} now becomes:
\begin{equation} \label{eq:2}
    p(b) \sim K \cdot P_{e}(b) \cdot D_{e}(p_{mc}) \cdot K^{-1} \cdot p_{mc}.
\end{equation}

From the aforementioned analysis, we conclude that we need to estimate depth $D_{e}$ at timestamp $t_{1}$ and the 6DoF pose change $P_{e}$ from $t_{1}$ to $t_{0}$, both from the input event representation $S_{e}$. To this end, we employ CNNs to model these quantities. We follow the traditional structure-from-motion pipeline \cite{zhou2017unsupervised} that utilizes an encoder-decoder to model depth, and a separate encoder to model pose. 

\paragraph{Baseline Architecture.} All of our networks follow the same baseline architecture, i.e. a ResNet34 encoder \cite{he2016deep} followed by a U-Net \cite{ronneberger2015u} decoder with skip connections going from the encoder features to the corresponding same-resolution decoder features. We want to emphasize that the proposed pipeline is architecture-agnostic, and as such, more advanced network architectures can be used to further improve the performance. However, the latter is beyond the scope of this paper, and in practice a ResNet34 encoder with U-Net decoder provides a good trade-off between performance and runtime. 

\paragraph{Depth and pose networks.} Our depth network uses the baseline architecture described in the previous paragraph. At the end of the depth network a convolutional layer is appended that maps the output channels to 1-channel disparity values. In contrast, our pose network only utilizes the encoder part of the baseline architecture, and a head is appended at the end of the encoder consisting of a global average pooling layer, and a convolutional layer that maps the output channels to 6-channel pose values (3 for rotation, 3 for translation).

Using the predicted depth $D_{e}$ and pose $P_{e}$ estimates, we can use Equation \ref{eq:2} to ego-motion compensate the event representation $S_{e}$, ending up with $S_{e}^{mc}$ which is later used as input to our motion segmentation module.

\subsection{Optical Flow Estimation} \label{subsec:flow}

In parallel to our ego-motion compensation step, we estimate the optical flow $F_{e}$ from $t_{0}\rightarrow t_{1}$. Although initially, this might seem redundant, the intuition of this design choice is two-fold: First, by combining depth $D_{e}$ and 6DoF pose $P_{e}$ our ego-motion compensation module can output only rigid optical flow $F_{e}^{rig}$, i.e. optical flow caused by ego-motion, so complementing the latter with optical flow $F_{e}$ that accounts for both rigid and non-rigid motions provides more reasoning cues to our motion segmentation module. Second, optical flow has been shown to synergize with motion segmentation both in the event \cite{mitrokhin2018event} and RGB \cite{xie2022segmenting, yang2021learning, bideau2018moa, bideau2016s} camera domains, hence we also want to leverage such synergies in our pipeline. 

\paragraph{Optical Flow Network.} Our optical flow network uses the baseline architecture described in the previous section. At the end of the optical flow network, a convolutional layer is appended that maps the output channels to 2-channel spatial displacements $F_{e}$, in the x- and y-axis respectively.

\begin{figure}[]

\centerline{
  \includegraphics[width=\linewidth]{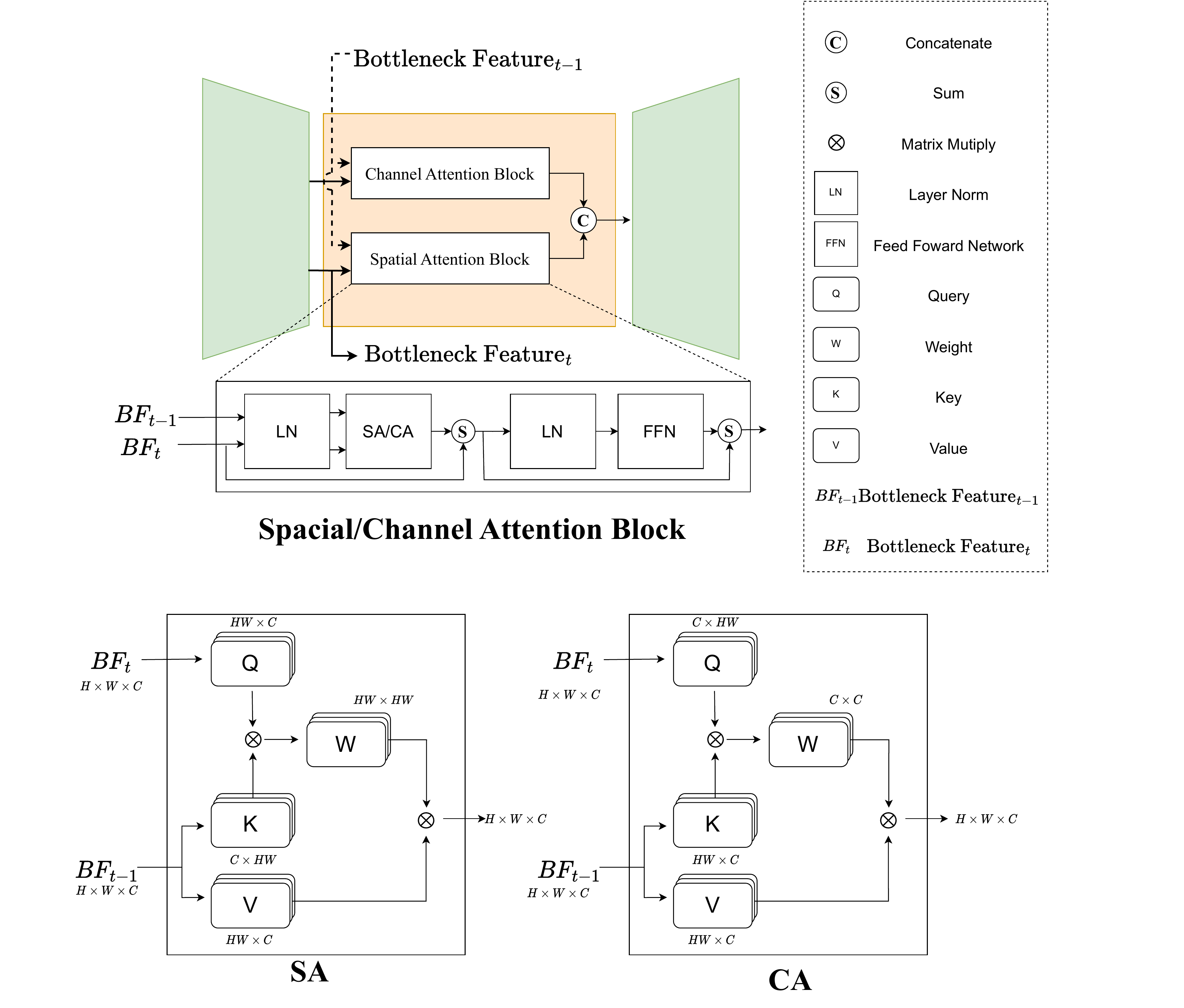}
}
  \caption{\textbf{Temporal Attention Module (TAM)}. The TAM that applies channel and spatial attention across the hidden states of different timestamps ($t$, $t-1$) is employed inside the motion segmentation network for temporally consistent motion masks.
  }
  \label{fig:TAM}
\end{figure}

\subsection{Motion Segmentation} \label{subsec:seg}

The ego-motion compensated event representation $S_{e}^{mc}$ and optical flow $F_{e}$ estimated in the previous steps are concatenated and used as input to the motion segmentation network that infers the segmentation mask $M$.  

\paragraph{Motion Segmentation Network.} It uses the same baseline architecture as before, which we enhance with a temporal attention module.

\paragraph{Temporal Attention Module.} In order to fix inconsistencies in temporally adjacent segmentation masks, typically caused by occlusions, noisy events, or insufficient events in-between time steps, we introduce a temporal attention module, placed at the bottleneck of the baseline architecture. Its architecture can be seen in Figure \ref{fig:TAM}. It consists of two (multi-head) query, key, value attention modules \cite{vaswani2017attention, zamir2022restormer} working in parallel, one applying the attention at the spacial level and the other at the channel level. To better understand their working principle let us assume that a hidden state, denoting the bottleneck features from a previous time step $BF_{t-1}$, is given as input to our motion segmentation network. We also extract the bottleneck features of the current time step $BF_{t}$. Consequently, the current time step bottleneck features (i.e. hidden state) $BF_{t}$ serve as the query, whereas the previous time step bottleneck features $BF_{t-1}$ are used as the key and value in our channel and spatial attention modules respectively. This effectively enables our motion segmentation network to select relevant information from the hidden state with a 'global receptive field' in mind. This is in stark contrast to the LSTM modules used in \cite{zhang2023multi}, which only have a 'local receptive field' or may require integration of multiple time steps before acquiring more global reasoning. The latter typically renders LSTMs rather expensive to train and prone to convergence instability. Finally, note that, during inference, the runtime of our network remains unaffected, and that we can accumulate information from multiple previous time steps without re-training the temporal attention module.

\subsection{Training Losses}

Each model is trained with a range of losses. The motion compensation step sums the image reconstruction loss $L_{img}$, the edge aware smoothness loss $L_{sm}$, the contrast maximization loss $L_{cm}$ and, where data is available, the supervised losses $L_{sup}$. Their respective weights are denoted by $\lambda_{*}$. Consequently, the motion compensation loss is:

\begin{equation}
    L_{MC} = \lambda_{img} L_{img} + \lambda_{sm} L_{sm} + \lambda_{cm} L_{cm} + \lambda_{sup} L_{sup}
\end{equation}

Optical flow is trained using the image reconstruction loss $L_{img}$, the smoothness loss $L_{reg}$, and the motion cycle consistency loss $L_{cyc}$:

\begin{equation}
    L_{OF} = \lambda_{img} L_{img} + \lambda_{reg} L_{reg} + \lambda_{cyc} L_{cyc}
\end{equation}

Finally, motion segmentation is supervised using the binary cross entropy loss $L_{bce}$. The exact definitions of each loss as well as their respective weights are provided in the supplementary material.

\section{Experiments}

In this section, we first introduce our DSEC-based dataset for motion segmentation and tracking, i.e. DSEC-MOTS, and then present quantitative and qualitative motion segmentation results on the DSEC-MOTS and EV-IMO benchmarks. For additional experimental details and results, we invite the reader to visit the supplementary material.

\subsection{DSEC-MOTS}

We note the lack of a large-scale real-world event dataset with motion segmentation masks of dynamic objects in complex scenes with arbitrary objects and motion. To remedy this situation, we introduce our such dataset, termed DSEC-MOTS, annotated based on the largest real-world autonomous driving event dataset, namely DSEC. Note that, \citet{sun2022ess} proposed DSEC-Semantic, which only includes semantic object annotations for both static and dynamic objects. A concurrent work DSEC-MOD \cite{zhou2022rgb} added bounding boxes for dynamic object detection, but only used a subset of DSEC where more than three dynamic objects exist. Most recently, DSEC-MOS \cite{zhou2023dsec} additionally extends DSEC-MOD with motion segmentation masks. However, at the time of this submission the dataset has not yet been publicly released. Different from DSEC-MOD and DSEC-MOS, we keep all sequences with good event quality, even when there are no dynamic objects since we believe such a choice is more general for class-agnostic motion segmentation. Furthermore, DSEC-MOTS also includes tracking of the motion segmentation masks. The supplementary material outlines in detail how DSEC-MOTS is build from the original DSEC dataset. The dataset is available at \url{https://github.com/rwn17/DSEC-MOTS}.

\begin{table}
\centering
\caption{\textbf{Comparison of baseline methods on our proposed DSEC-MOTS benchmark}. IoU refers to the IoU values of moving objects, as in traditional 2D segmentation metric.}
\begin{tabular}{llll}
\hline
& EV-IMO \cite{mitrokhin2019ev} & EV-IMO w/ ECN \cite{ye2020unsupervised} & Ours  \\
\hline
IoU(\%) & 31.61  & 39.95 & \textbf{52.86} \\
\hline
\label{tab:dsec-dos}
\end{tabular}
\end{table}

\begin{figure*}
\includegraphics[width=\textwidth]{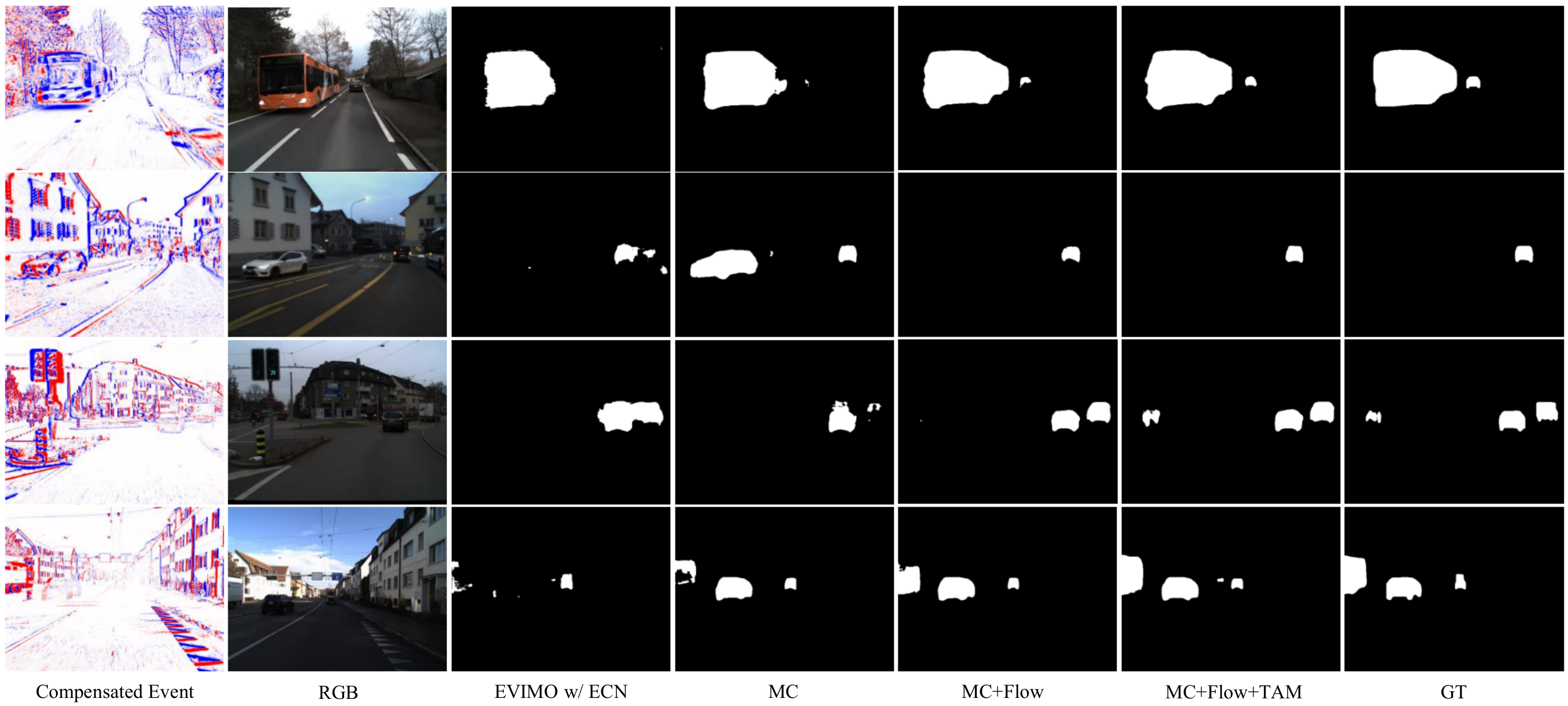}
\caption{\textbf{Comparative qualitative analysis of the baseline model and variations of our model.} From left to right: ego-motion compensated events, RGB image (only for reference),  EV-IMO \cite{mitrokhin2019ev} with ECN backbone \cite{ye2020unsupervised} segmentation mask, our segmentation mask (only ego-motion compensation), our segmentation mask (ego-motion compensation plus optical flow), our segmentation mask (full model), and ground truth motion segmentation mask.}
\label{fig:dsec-ablation}
\end{figure*}

\subsection{Motion Segmentation on DSEC-MOTS}

In this section, we compare our method against the EV-IMO \cite{mitrokhin2019ev}, and EV-IMO with ECN backbone \cite{ye2020unsupervised} CNNs on the DSEC-MOTS dataset. For a fair comparison, we only replace our networks with the EV-IMO ones, and train them using the exact same losses, settings, and protocol. This way EV-IMO models can further benefit from our more elaborate training scheme and losses. As the comparison metric, we choose the standard Intersection of Union (IoU) on the dynamic objects.

Note that, the aforementioned approaches are the only CNN-based ones with publicly available code required to train them from scratch on DSEC-MOTS. The other learning-based works rely on SNNs \cite{parameshwara2021spikems} or GNNs \cite{mitrokhin2020learning} which have completely different training protocols that are not yet mature enough for datasets like DSEC-MOTS. Alternatively, optimization-based works \cite{parameshwara20210, he2021fast, zhou2021event, falanga2020dynamic, stoffregen2019event, mitrokhin2018event} require cumbersome hyper-parameter tuning, with no guarantee that they can work in a scenario (i.e. autonomous driving scenes) so different from the one they were initially designed for (i.e. indoor scenes with toy dynamic objects). Still, for completeness in the following section we provide a comparison with these methods, including \cite{zhang2023multi}, on the EV-IMO benchmark. 

Table \ref{tab:dsec-dos} summarizes the results on DSEC-MOTS. Our method improves significantly over the next best one, i.e. 12.91 IoU improvement. This is expected as our baseline architecture described in Section \ref{subsec:mot_comp} has stronger reasoning capability than that of EV-IMO models which were designed with efficiency also in mind. Yet, looking at the ablation study in Table \ref{tab:dsec-ablation} we can clearly see that even somehow upgrading the EV-IMO models to the performance of our baseline architecture, i.e. 51.02 IoU, we still manage to improve another 1.84 IoU by adding the proposed motion compensation, optical flow, and temporal attention parts respectively. Figure \ref{fig:dsec-ablation} shows qualitative examples of how our method improves over the compared works.

\begin{table}[]
\centering
\caption{\textbf{Comparison with the state-of-the-art on the EV-IMO benchmark}. pIoU(\%) is the IoU of moving objects in points, as defined in the SpikeMS paper \cite{parameshwara2021spikems}, which is the default metric of choice in EV-IMO. IoU(\%) is the IoU for moving object, as in traditional 2D segmentation IoU. And mIoU(\%) is the mean IoU of moving objects and the background. }
\begin{tabular}{lccc}
\toprule
& pIoU & IoU & mIoU \\
\midrule
EVDodgeNet\cite{sanket2020evdodgenet} & 65.76 &    &   \\
EV-IMO\cite{mitrokhin2019ev}          & 77.00 &    &   \\ 
0-MMS \cite{parameshwara20210}        & 80.37 &    &   \\
MSRNN\cite{zhang2023multi}            &       & 65.9   & 81.0  \\
Ours                                  &\textbf{ 84.89 }  & \textbf{68.1}   & \textbf{83.2}  \\
\bottomrule
\end{tabular}
\label{tab:evimo_seg}
\end{table}

\begin{table}[]

\centering
\caption{\textbf{Comparison with the state-of-the-art on different backgrounds of the EV-IMO benchmark.} Here, pIoU(\%) is the IoU of moving objects in points.}
\begin{tabular}{llllll}
\hline
\multicolumn{1}{c}{\multirow{2}{*}{Method}} & \multicolumn{5}{c}{Backgrounds}  \\
\multicolumn{1}{c}{}  
& \multicolumn{1}{c}{boxes} & \multicolumn{1}{c}{floor} & \multicolumn{1}{c}{wall} & \multicolumn{1}{c}{table} & \multicolumn{1}{c}{fast} \\
\hline
EV-IMO\cite{mitrokhin2019ev}                 & 70 & 59 & 78 & 79 & 67 \\
EVDodgeNet\cite{sanket2020evdodgenet}        & 67 & 61 & 72 & 70 & 60 \\
SpikeMS \cite{parameshwara2021spikems}       & 65 & 63 & 63 & 50 & 38 \\
GConv \cite{mitrokhin2020learning}           & 60 & 55 & 80 & 51 & 39 \\
Ours                                         & \textbf{77} & \textbf{94} & \textbf{85} & \textbf{87} & \textbf{80} \\
\hline
\end{tabular}
\label{tab:evimo_bgs}
\end{table}

\begin{figure}[]
\centerline{
\includegraphics[width=\linewidth]{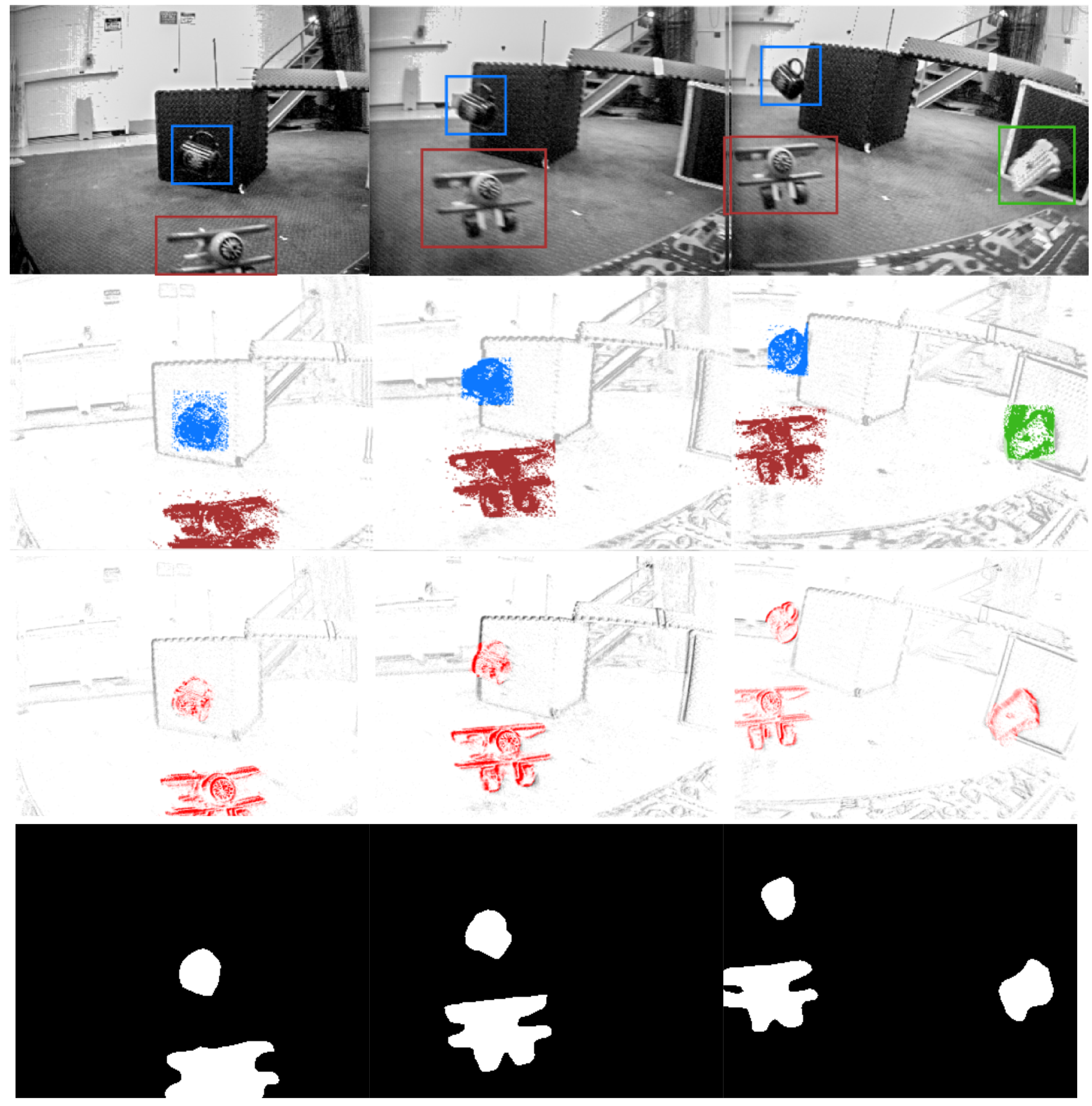}
}
\caption{\textbf{Qualitative comparison between our algorithm and 0-MMS \cite{parameshwara20210} on the EV-IMO benchmark.} From top to bottom: RGB image, 0-MMS event segmentation, our event segmentation, and our segmentation mask. The first two rows are taken from the 0-MMS \cite{parameshwara20210} paper.}
\label{fig:evimo}
\end{figure}

\subsection{Motion Segmentation on EV-IMO}

Although our main focus is motion segmentation for challenging outdoor scenes, our method sets the new state-of-the-art on the EV-IMO benchmark, still the most prominent dataset for event-based motion segmentation, with large improvements across a variety of different methods and metrics. We follow the established evaluation protocol \cite{mitrokhin2019ev, parameshwara2021spikems} and use as metric the IoU of moving objects in points (referred to as pIoU), as well as the traditional dense IoU for moving objects (referred to as IoU) and the mean IoU of moving objects and background (referred to as mIoU) as in MSRNN\cite{zhang2023multi}. Table \ref{tab:evimo_seg} quantitatively shows that our method achieves large improvements of 2.19 IoU (2.22 mIoU) over MSRNN \cite{zhang2023multi}, and 4.52 pIoU over 0-MMS \cite{parameshwara20210}, the state-of-the-art methods in each metric respectively. Furthermore, Table \ref{tab:evimo_bgs} shows the performance of different approaches on the EV-IMO benchmark with respect to different backgrounds, a common practice in this benchmark. We observe that regardless of the background our method consistently outperforms its counterparts. Besides the quantitative improvements, observing Figure \ref{fig:evimo} we can see that our method has sharper object boundaries compared to 0-MMS. As 0-MMS is using a clustering-based method, it inevitably gets false positive segmentation results around objects. In contrast, our algorithm can predict a sharper boundary for the moving objects. 

\begin{table}[]
\caption{\textbf{Ablation study on the DSEC-MOTS and EV-IMO benchmarks} Comparing proposed motion compensation (MC), optical flow (OF), and temporal attention (TAM) steps. Here, IoU refers to the IoU of moving objects, and the numbers in brackets show the mean IoU, the average of moving objects and background, as in MSRNN \cite{zhang2023multi}.}
\centering
\begin{tabular}{lllcc}
\hline
MC & OF & TAM & DSEC-MOTS &EVIMO  \\
\hline
           &            &             & 51.02 (75.21)  & 65.95 (81.60) \\
\checkmark &            &             & 51.48 (75.46)  & 67.68 (82.57)\\
\checkmark & \checkmark &             & 52.02 (75.71)  & 67.71 (82.60)\\
\checkmark & \checkmark & \checkmark  & \textbf{52.86} (\textbf{76.16}) &\textbf{ 68.09} (\textbf{83.22}) \\
\hline\\
\label{tab:dsec-ablation}
\end{tabular}
\end{table}

\subsection{Ablation Study}

In this section, we quantitatively and qualitatively ablate the contribution of each proposed step in our pipeline, i.e. motion compensation (MC), optical flow estimation (OF), and the temporal attention module (TAM). From Table \label{tab:dsec-ablation} we observe that on DSEC-MOTS adding one by one step to our baseline architecture brings consistent improvements of 0.46, 0.54, and 0.84 in IoU of moving objects respectively. Similar observations can be made on EV-IMO where each step brings improvements of 1.73, 0.03, and 0.38 in IoU of moving objects respectively. Another useful observation is that different datasets tend to benefit more from the step that directly accounts for their deficiency. For example, on DSEC-MOTS due to excessive noise and jittering in the events the TAM step brings the most improvement. In contrast, on EV-IMO due to large camera motions within a few milliseconds the MC step is by far the most important step.

Figure \ref{fig:dsec-ablation} illustrates qualitatively what each step adds to the final result. We can see that the segmentation results from raw event input (columns 3 and 4) tend to have more noisy boundaries and floating parts compared to their ego-motion compensated (MC) counterparts. The optical flow (OF) input can help to discover objects as well as reject false positive segmentation, like a still car. Temporal attention (TAM) can help to maintain temporal consistency against the jittery nature of event cameras, especially when the moving objects have the same speed as ego motion. Also, temporal attention (TAM) can help to improve the segmentation results on large moving objects, whose middle part triggered a few events due to the lack of texture. More such examples can be found in the supplementary material.




\section{Conclusion} 
\label{sec:conclusion}

In this work, we push event-based motion segmentation from small-scale floating toy scenes to complex large-scale scenes, like autonomous driving. We present, a class-agnostic event-based motion segmentation pipeline, together with DSEC-MOTS, a moving object segmentation dataset based on the large-scale dataset DSEC. Our pipeline utilizes ego-motion compensation and optical flow cues as well as temporal information to improve beyond previous state-of-the-art algorithms by a large margin. We evaluate our algorithm through indoor and complex outdoor datasets.

{
    \small
    \bibliographystyle{ieeenat_fullname}
    \bibliography{references}
}

\end{document}


\maketitlesupplementary

\section{Training losses}

In this section we describe the training losses of our pipeline. Let us assume the following variables: (i) the predicted depth $D_{e}$, 6DoF pose $P_{e}$, optical flow $F_{e}$, and segmentation mask $M$, as well as the rigid optical flow $F_{e}^{rig}$ computed from $D_{e}$ and $P_{e}$; (ii) optionally sparse (across both pixels and samples) ground truth about depth $\hat{D}_{e}$ and rigid optical flow $\hat{F}_{e}^{rig}$, as well as dense ground truth for the segmentation masks $\hat{M}$; (iii) a pair of RGB images $(I_{bwd}, I_{fwd})$ spatially aligned with the event representation $S_{e}$ and temporally aligned with the beginning (i.e. $I_{bwd}$) and end (i.e. $I_{fwd}$) of the event representation respectively, as well as their warped counterparts $(\hat{I}_{bwd} = F_{Warp}(I_{fwd}; D_{e}^{bwd}; -P_{e}), \hat{I}_{fwd} = F_{Warp}(I_{bwd}; D_{e}^{fwd}; P_{e}))$ generated using a differentiable inverse warping function $F_{warp}$ \cite{zhou2017unsupervised, jaderberg2015spatial} whose source sampling locations are found using a slight variation of Equation 2 from the main paper: 
\begin{equation} \label{eq:2}
    p(b) \sim K \cdot P_{e}(b) \cdot D_{e}(p_{mc}) \cdot K^{-1} \cdot p_{mc}
\end{equation}
for the RGB case\footnote{Note that, $D_{e}^{fwd} \equiv D_{e}$ here, and $D_{e}^{bwd}$ is calculated by reversing the raw asynchronous event stream \cite{tulyakov2021time} to get the reversed event representation $S_{e}^{rev}$, and inputting the latter to the depth network.}; (iv) the ego-motion compensated event representation $S_{e}^{mc}$ generated using $F_{warp}$ and Equation \ref{eq:2}.

\textbf{Motion compensation losses.} The total loss of the motion compensation module is comprised of the following losses:  
\begin{equation} \label{eq:mc}
    L_{MC} = \lambda_{img} L_{img} + \lambda_{sm} L_{sm} + \lambda_{cm} L_{cm} + \lambda_{sup} L_{sup}.
\end{equation}

First, $L_{img}$ is the image reconstruction loss applied between pairs of original and warped images respectively, i.e. $(I_{*}, \hat{I}_{*})$ with $*$ denoting $bwd$ or $fwd$, and given by:
\begin{equation} \label{eq:3}
\begin{split}
    L_{img} = \frac{1}{\vert V \vert} \sum_{p \in V} ( & 0.15 \cdot \|I_{*}(p) - \hat{I}_{*}(p)\|_{1} + \\ 
    & 0.85 \cdot \frac{1 - SSIM(I_{*}(p), \hat{I}_{*}(p))}{2} ),
\end{split}
\end{equation}
with $SSIM$ being the widely-used metric to measure image similarity \cite{wang2004image}, and $V$ standing for the set of valid pixels with invalid ones accounting for out-of-image projections, depth inconsistent occlusions \cite{bian2021unsupervised}, and stationary points \cite{godard2019digging}; 

Second, $L_{sm}$ is an edge-aware smoothness loss to regularize the predicted depth maps where there are no events, given by:
\begin{equation}
    L_{sm} = \sum_{p} ( e^{-\nabla I_{*}(p)} \cdot \nabla D_{e}^{*}(p) )^{2},
\end{equation}
with $\nabla$ being the first derivative along spatial dimensions.

Third, $L_{cm}$ is the contrast maximization loss \cite{shiba2022secrets, gallego2018unifying}, measuring how much the contrast increased in the ego-motion compensated event representation $S_{e}^{mc}$ relative to the input event representation $S_{e}$, and given by:
\begin{equation}
    L_{cm} = \frac{F_{cm}(S_{e}^{mc})}{F_{cm}(S_{e}) + \epsilon}, \text{~with~} F_{cm}(S) = \frac{1}{\vert V \vert} \sum_{p \in V} \| \nabla S(p) \|^{2},
\end{equation}
where similarly to \cite{shiba2022secrets} we calculate this loss at multiple timestamps ($mc \in \{bwd, fwd\}$ here).

Finally, $L_{sup}$ denotes the supervised losses, when ever (i.e. samples) and where ever (i.e. pixels) they are available, and is given by:
\begin{equation}
\begin{split}
    L_{sup} = & \frac{1}{\vert A_{D} \vert} \sum_{p \in A_{D}} \|D_{e}(p) - \hat{D}_{e}(p)\|_{1} + \\
    & \frac{1}{\vert A_{F} \vert} \sum_{p \in A_{F}} \|F_{e}^{rig}(p) - \hat{F}_{e}^{rig}(p)\|_{1},
\end{split}
\end{equation}
where $A_{*}$ is the set of pixels with available ground truth for the corresponding variable ($D_{e}$ and $F_{e}^{rig}$ here). 

\textbf{Optical flow losses.} The total loss of the optical flow module is comprised of the following losses:  
\begin{equation} \label{eq:of}
    L_{OF} = \lambda_{img} L_{img} + \lambda_{reg} L_{reg} + \lambda_{cyc} L_{cyc}.
\end{equation}

First, $L_{img}$ is the image reconstruction loss defined in Equation \ref{eq:3}, applied between pairs of original and warped images respectively, i.e. $(I_{*}, \hat{I}_{*})$ with $*$ denoting $bwd$ or $fwd$. Here, $(\hat{I}_{bwd} = F_{Warp}(I_{fwd}; F_{e}^{bwd}), \hat{I}_{fwd} = F_{Warp}(I_{bwd}; F_{e}^{fwd}))$ are generated using the differentiable inverse warping function $F_{warp}$ \cite{jaderberg2015spatial}, but the source sampling locations are found directly from the predicted optical flow\footnote{Note that, $F_{e}^{fwd} \equiv F_{e}$ here, and $F_{e}^{bwd}$ is calculated by reversing the raw asynchronous event stream \cite{tulyakov2021time} to get the reversed event representation $S_{e}^{rev}$, and inputting the latter to the optical flow network.}. In contrast to the motion compensation image reconstruction loss, the set of invalid pixels here accounts only for out-of-image projections and occlusions, as the optical flow can account for dynamic objects.

Second, $L_{reg}$ is a smoothness loss to regularize the predicted optical flow $F_{e}$ where there are no events, given by:
\begin{equation}
    L_{reg} = \sum_{p} ( \nabla F_{e}(p) )^{2}.
\end{equation}

Finally, $L_{cyc}$ is a motion cycle consistency loss that enforces the backward $F_{e}^{bwd}$ and forward $F_{e}^{fwd}$ optical flows to be opposite of each other and is given by:
\begin{equation}
    L_{cyc} = \sum_{p} \frac{\| F_{e}^{bwd} + F_{Warp}(F_{e}^{fwd}; F_{e}^{bwd}) \|^{2}}{\| F_{e}^{bwd} \|^{2} + \| F_{Warp}(F_{e}^{fwd}; F_{e}^{bwd}) \|^{2} + \epsilon}.
\end{equation}
Note that, in practice, this loss is applied twice, once for forward cycle consistency and once for backward cycle consistency, with the latter achieved just by switching $fwd$ and $bwd$ notations in the above equation. 

\textbf{Motion segmentation losses.} For this part we use the binary cross entropy loss between predicted $M$ and ground truth $\hat{M}$ segmentation masks:
\begin{equation} \label{eq:ms}
    L_{bce} = \sum{p} -(\hat{M} \cdot \log (M) + (1 - \hat{M}) \cdot \log (1 - M)),
\end{equation}
where we use ten times higher weight for the foreground class (i.e. dynamic objects) versus the background class (i.e. static objects and background), as the latter appears much more frequently with respect to the former. 

Normally, when a multitude of losses is considered, as is the case for motion compensation and optical flow modules, the weighting factors $\lambda_{*}$ of each loss should be balanced. However, in practice we found that converting all variables in the pixel range (i.e. disparity as pixel x-displacement, optical flow as pixel xy-displacements, image reconstructions as pixel differences, etc.) allowed us to simply set all $\lambda_{*} = 1$.

\section{Implementation Details}

\paragraph{DSEC-MOTS Annotation Details}In order to generate segmentation masks for dynamic objects, we first apply a state-of-the-art LiDAR-based dynamic object segmentation algorithm 4DMOS \cite{mersch2022ral}, and then project the LiDAR points to the RGB image plane. Consequently, we run a tracker \cite{pcan} on the RGB images to get their unique ID as well as temporally consistent instance masks. An instance mask is marked as dynamic if the projected LiDAR points hit more than three RGB frames. The instance masks of dynamic objects are then projected to the event plane, based on planar-depth assumption as done in the original DSEC dataset. As a final step, we manually correct the wrong masks. We annotate all the DSEC training sequences (41 sequences, 28179 images) and test sequences (10 sequences, 3956 images) following DSEC separation for the left event camera.

\paragraph{Training Details}We train our pipeline in stages. All training is performed on the DSEC and DSEC-MOTS datasets. Since DSEC dataset only contains train and test splits, we also generate a validation split by removing five sequences from the train set, i.e. interlaken\_00\_e, zurich\_city\_01\_e, zurich\_city\_02\_e, zurich\_city\_04\_e, and zurich\_city\_09\_e. Consequently, we train all networks on the train set, select the best-performing checkpoint on the validation set, and deploy it on the test set.

First, we pre-train the motion compensation module, meaning the depth and pose networks. We use the losses described in Equation \ref{eq:mc}, all with equal weights $\lambda_{*} = 1$. We use a batch size of 8, which effectively translates to 16 as we use both the normal and inverted event sequence as described in the paper to predict depth at timestamps $t_{1}$ and $t_{0}$ and pose $t_{1} \rightarrow t_{0}$ respectively (for the pose $t_{0} \rightarrow t_{1}$ we simply invert the pose $t_{1} \rightarrow t_{0}$). We use Adam optimizer with learning rate $1^{-4}$ and train for 50 epochs. We use random scale uniformly sampled in $[0.80, 1.25]$, random crop of size $512 \times 384$, and random horizontal flip with $0.50$ probability. Note that, we deactivate all augmentations during the last 5 epochs.

Second, we pre-train the optical flow module, meaning the optical flow network. We use the losses described in Equation \ref{eq:of} of the paper, all with equal weights $\lambda_{*} = 1$. We use a batch size of 8, which effectively translates to 16 as we use both the normal and inverted event sequence as described in the paper to predict the optical flow $t_{1} \rightarrow t_{0}$ and the optical flow $t_{0} \rightarrow t_{1}$ respectively. The remaining training details are identical to the motion compensation module.

Finally, we freeze the pre-trained motion compensation and optical flow modules, and train the motion segmentation module. We use the loss described in Equation \ref{eq:ms} of the paper. We use a batch size of 8, which effectively translates to 16 as we use both the normal and inverted event sequence as described in the paper to predict the motion segmentation masks at $t_{1}$ and $t_{0}$ respectively. The remaining training details are identical to the previous modules. 

For the EV-IMO training, due to the lack of background depth, we choose a homography estimation network instead of our previous pose-depth network for ego-motion compensation with the exact same architecture as the depth network that regresses xy-displacements of the four image corners (top-left, top-right, bottom-right, bottom-left). We find that homography can model the motion in the EV-IMO dataset quite well, since most of the motions are rotational.

\begin{figure*}
  \includegraphics[width=\textwidth]{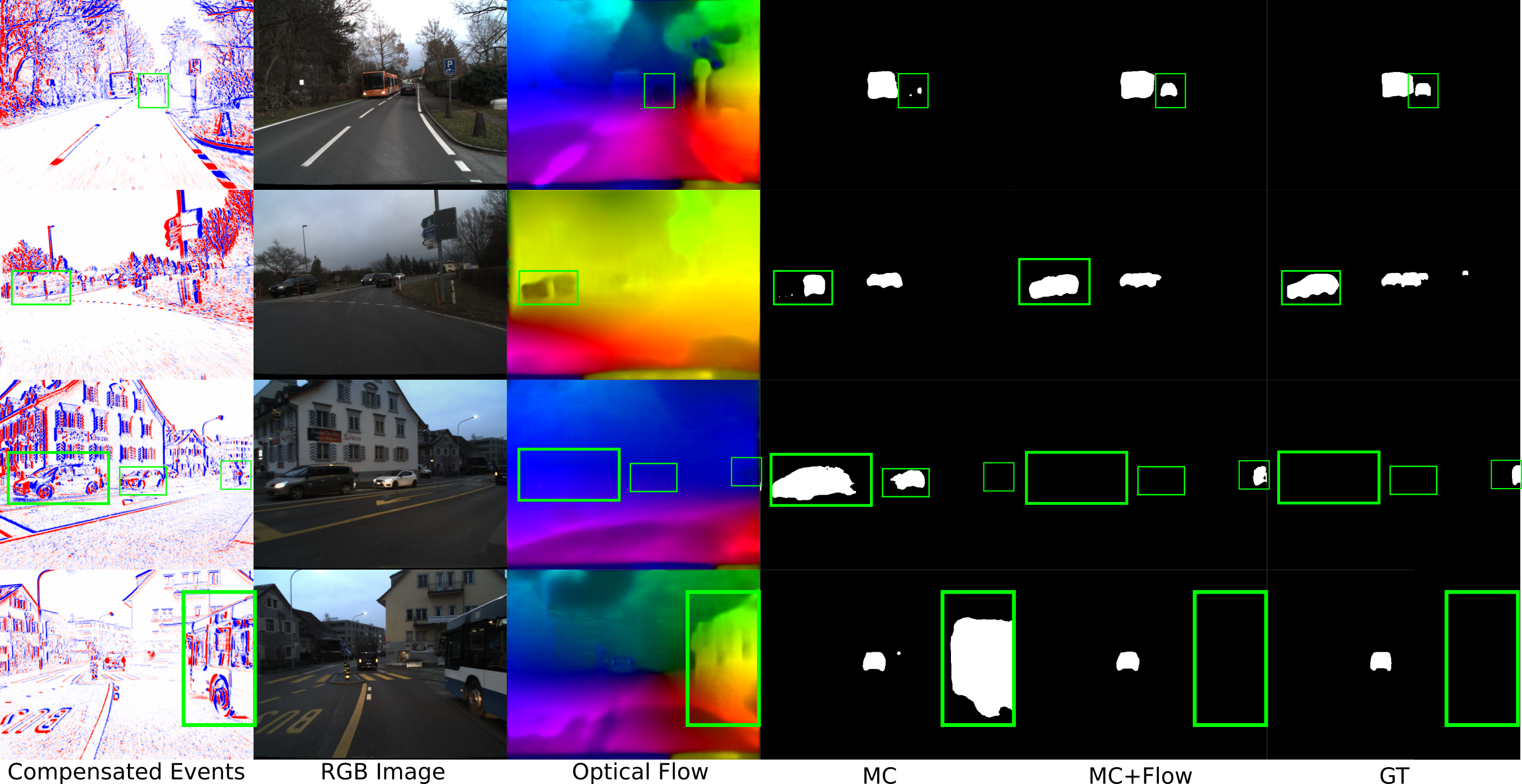}
  \caption{Qualitative results of motion segmentation on DSEC-MOTS (part 1). From left to right: event representation, RGB image, predicted optical flow, our approach with the motion compensation step, our approach with the motion compensation and optical flow steps, and the ground truth. The comparison is highlighted by the green boxes.
  }
  \label{fig:supp_flow}
\end{figure*}

\begin{figure*}
  \includegraphics[width=\textwidth]{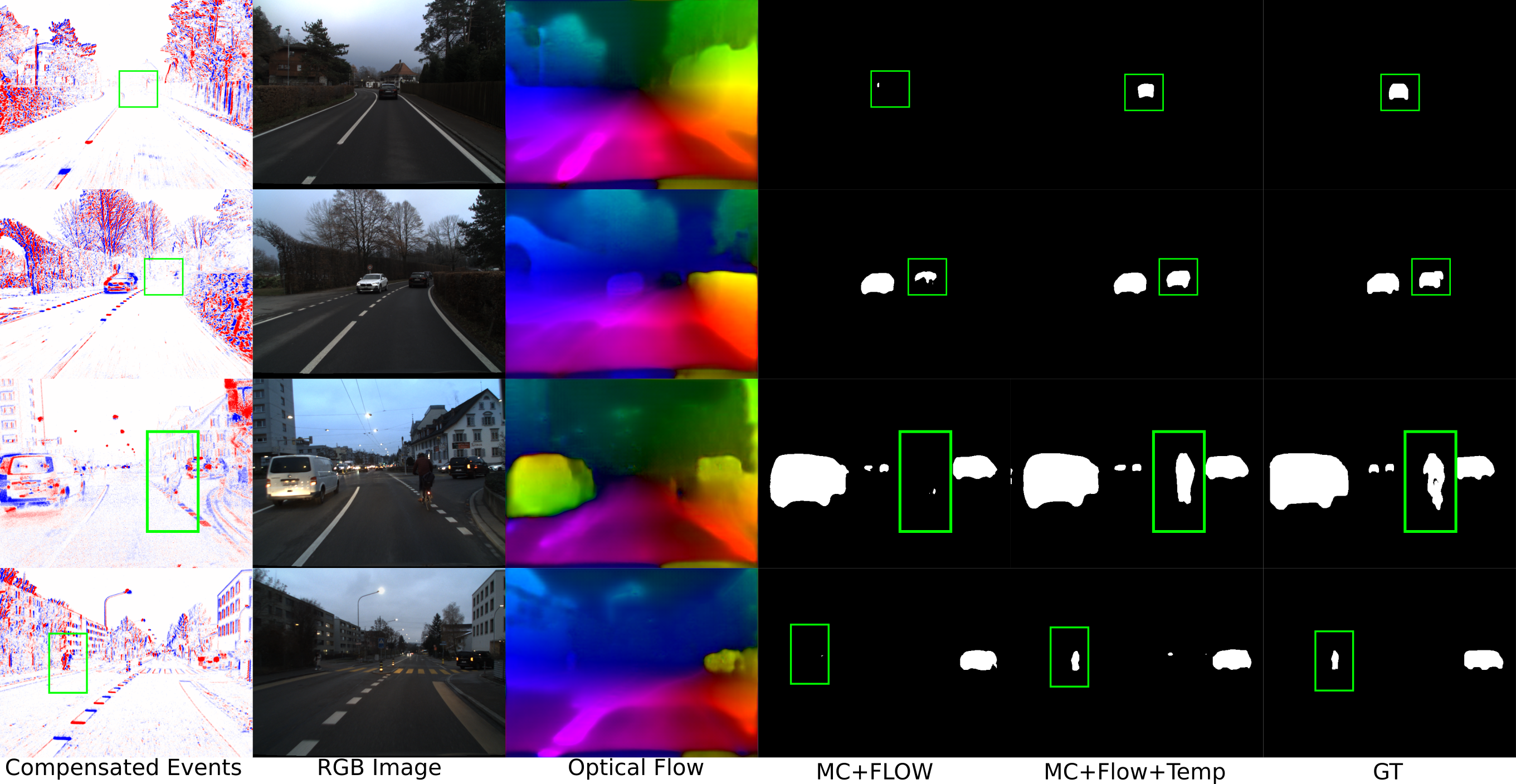}
  \caption{Qualitative results of motion segmentation on DSEC-MOTS (part 2). From left to right: event representation, RGB image, predicted optical flow, our approach with the motion compensation and optical flow steps, our approach with the motion compensation, optical flow and temporal consistency steps, and the ground truth. The comparison is highlighted by the green boxes.
  }
  \label{fig:supp_temp}
\end{figure*}

\section{Extra Results on Motion Segmentation}

Additional results from our ablation study on motion segmentation are available in Figures \ref{fig:supp_flow} and \ref{fig:supp_temp}. 

Figure \ref{fig:supp_flow} shows that when we add the optical flow step, the network can now recognize objects that it previously missed. For example, in the first row, it can detect a car moving at the same speed as the camera. In the next row, thanks to the optical flow, the network can predict a more complete segmentation mask. This feature also helps reduce mistakes, as seen in the third and fourth rows. Figure \ref{fig:supp_temp} provides more examples of how our motion segmentation pipeline produces clearer and more consistent masks, especially given the shaky nature of the event camera in an autonomous driving setting.


\begin{figure*}
  \includegraphics[width=\textwidth]{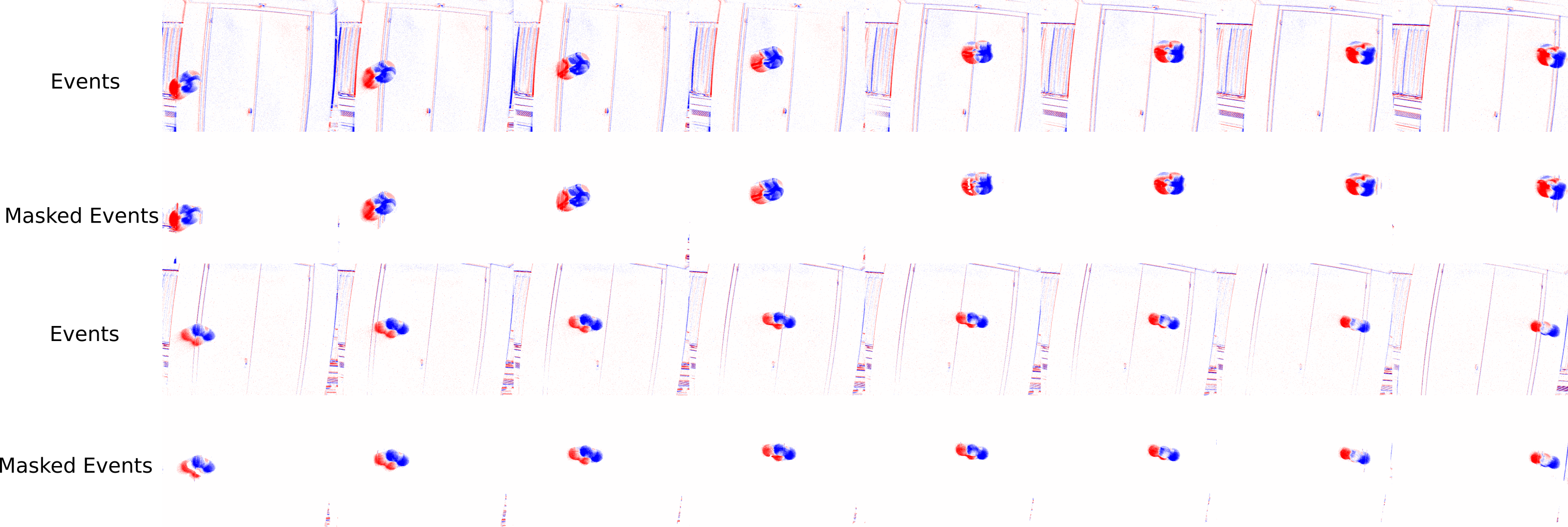}
  \caption{Qualitative results of motion segmentation on a manually collected indoor dataset without any retraining. The first two rows show a sequence with flying volleyball and the last two rows a sequence with flying thread ball. No ground truth motion segmentation masks are available here.
  }
  \label{fig:supp_indoor}
\end{figure*}

\section{Generalization to Unseen Datasets}

\begin{figure*}
\begin{minipage}[c][\textheight]{\textwidth}
\centering
\includegraphics[width=0.95\textwidth]{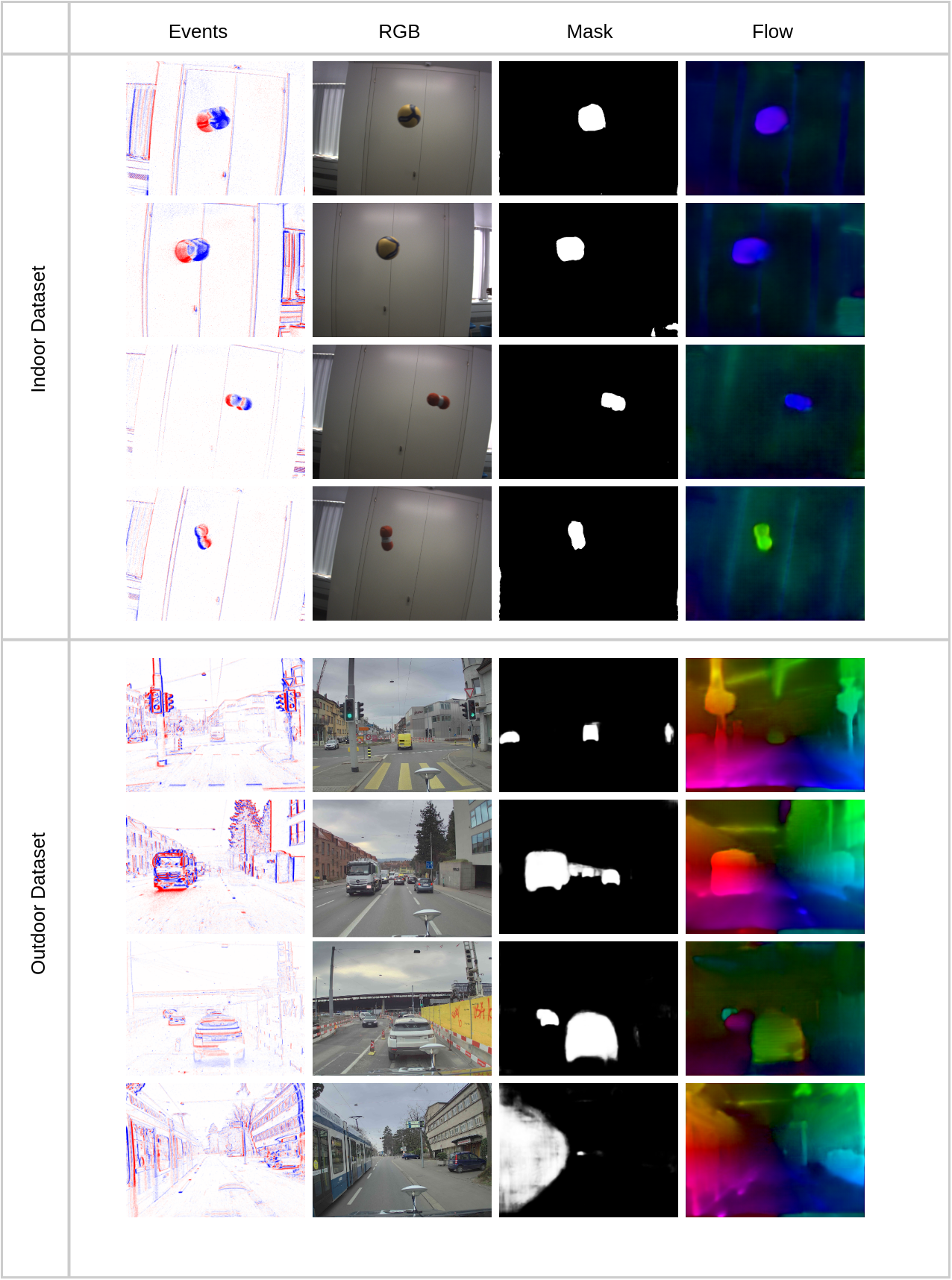}
\caption{Qualitative results of motion segmentation masks and optical flow fields on unseen datasets. Our model is only trained on DSEC-MOTS and deployed on unseen datasets without additional fine-tuning.}
\label{fig:gene}
\end{minipage}
\end{figure*}

We also test the generalization capabilities of our algorithm on unseen datasets, as shown in Figure \ref{fig:gene}. 

We find that our algorithm trained on autonomous driving scenes can be directly deployed on real-world indoor scenes with unseen objects that we recorded ourselves, without any fine-tuning (see Figure \ref{fig:supp_indoor}). This shows that our algorithm can capture the true nature of motion instead of just over-fitting certain classes on autonomous driving scenes (e.g. cars, pedestrians, etc.). This class-agnostic aspect of our algorithm can be potentially useful in other tasks, like robot emergency collision avoidance. This is also one of the reasons that we model it as a segmentation task, instead of a detection task. 

The same holds when our method is deployed to real-world autonomous driving scenes, different from DSEC, that we recorded ourselves (see Section \ref{sec:robo}). Our algorithm can transfer to a new dataset with an event camera with different intrinsic and different baseline setups. It works well for crowded driving scenes (row two) as well as challenging scenes where a tram passes by (row four). We also provide a video for the generalization experiments.


\section{Downstream Robotic Applications} \label{subsec:application}

We demonstrate potential benefits of our proposed pipeline in robotic downstream applications. Specifically, we demonstrate the following tasks:
\begin{enumerate}
    \item Sampling and tracking of key-points for ego-motion estimation
    \item Pose estimation in dynamic environments in combination with LiDAR odometry and mapping
    \item Drivable Surface Detection
\end{enumerate}
For evaluations, we use the DSEC sequences with length $> 25s$ and the self-collected driving data (see Section \ref{sec:robo}).

\begin{table}[]
\caption{Keypoint tracking statistics}
\centering
\begin{tabular}{lll}
\hline
          & Acc.  & Recall \\
\hline
Zurich\_city\_05\_a    & 66 & 43  \\
Zurich\_city\_13\_b      & 69 & 53 \\
Interlaken\_00\_b & 73 & 68 \\
Zurich\_city\_13\_a      & 86 & 82 \\
Interlaken\_00\_a      & 84 & 86  \\
\hline
\end{tabular}
\label{tab:keypointtracking}
\end{table}

\begin{figure}[]
\centerline{
\includegraphics[width=\linewidth]{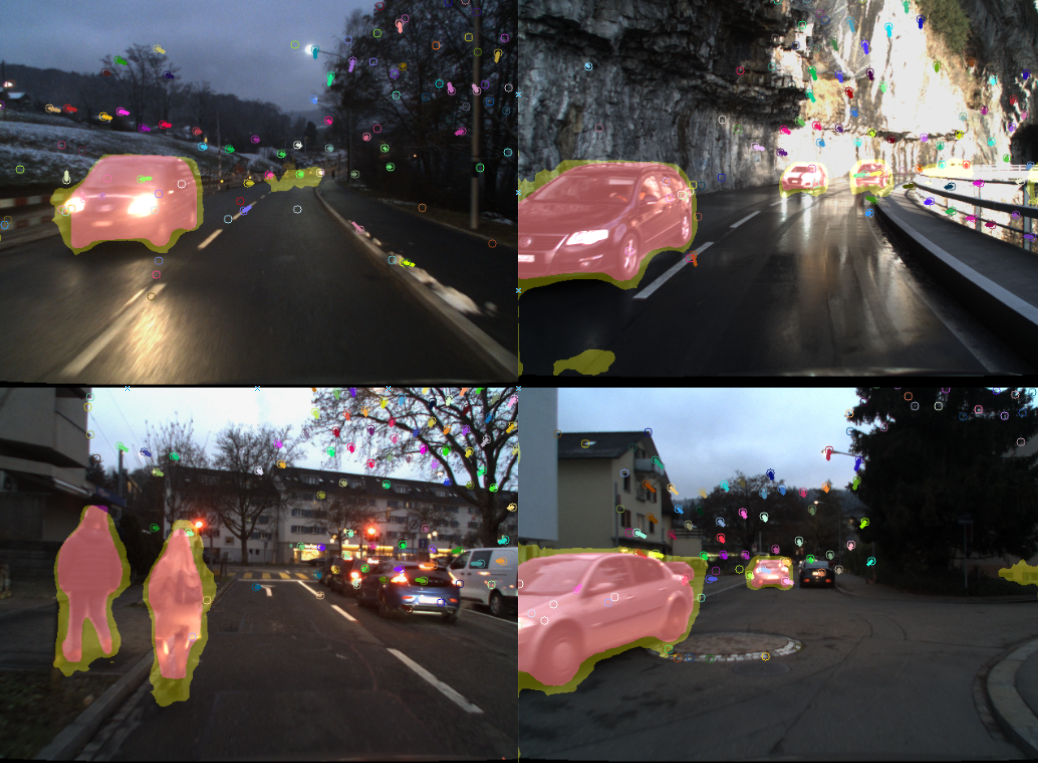}
}
\caption{Visualization of the ground truth (red) and inferred (yellow) masks overlayed on the image. Features tracks are shown before the filtering steps that will eliminate all the features falling within the yellow mask. }
\label{fig:feature_tracks}
\end{figure}

\subsection{Event-based key-point tracking}
In this experiment, we superimpose the proposed motion segmentation on key-point tracking, as commonly required for ego-motion estimation (see Fig.~\ref{fig:feature_tracks}). We apply the motion segmentation masks on a standard feature tracker~\cite{qin2018vins}, tracking 30 features, and evaluate recall and accuracy in a variety of sequences that ranges from low to high dynamic content on the DSEC-MOTS dataset (see Table~\ref{tab:keypointtracking}). 
In low dynamic content scenarios, usually corresponding to few and far away dynamic objects, the filter performance suffers slightly. SLAM algorithms are usually well-equipped to deal with these situations with simple outlier rejection and robust estimation techniques. Conversely, when in challenging scenarios, e.g. degradation through perceptual aliasing in combination with scene dynamics, the masks correctly filter up to 86\% of the tracks belonging to dynamic objects.  

\begin{table}[]
\caption{Event odometry with respect to LiDAR odometry~\cite{koide2021voxelized}}
\centering
\begin{tabular}{llll}
\hline
DSEC sequence & trans [\%] & rot [deg/100m] \\
\hline
interlaken\_00\_a & 4.78  & 0.31 \\
interlaken\_00\_b & 4.63  & 0.37 \\
interlaken\_01\_a & 4.34  & 0.40 \\
thun\_01\_b & 2.59  & 0.34 \\
zurich\_city\_01\_a & 3.43  & 0.17 \\
zurich\_city\_02\_d & 2.82  & 0.27 \\
zurich\_city\_02\_e & 3.54  & 0.197 \\
zurich\_city\_05\_a & 2.88  & 0.20 \\
zurich\_city\_12\_a & 4.04  & 0.37 \\
zurich\_city\_14\_b & 4.18  & 0.14 \\
zurich\_city\_14\_c & 3.85 & 0.52 \\
\hline
average & 3.73 & 0.29 \\
\hline
\end{tabular}
\label{tab:dsec}
\end{table}

\begin{table}[]
\caption{Event odometry and LiDAR odometry with respect to ground truth on own validation data}
\centering
\begin{tabular}{llll}
\hline
Sequence & trans [\%] events & trans [\%] LiDAR \\
\hline
2023-06-08-15-10-16 & 1.260  & 1.424 \\
2023-05-10-16-23-24 & 1.083  & 0.599 \\
2023-05-10-15-30-19 & 1.352  & 1.274 \\
2023-05-09-16-25-24 & 1.368 & 1.810 \\
2023-05-10-15-41-42 & 1.391 & 1.795 \\
2023-06-05-16-10-03 & 1.753 & 1.860 \\

\hline
average & 1.368 & 1.460 \\
\hline
\end{tabular}
\label{tab:robo}
\end{table}

\begin{figure}[]
\centerline{
\includegraphics[width=\linewidth]{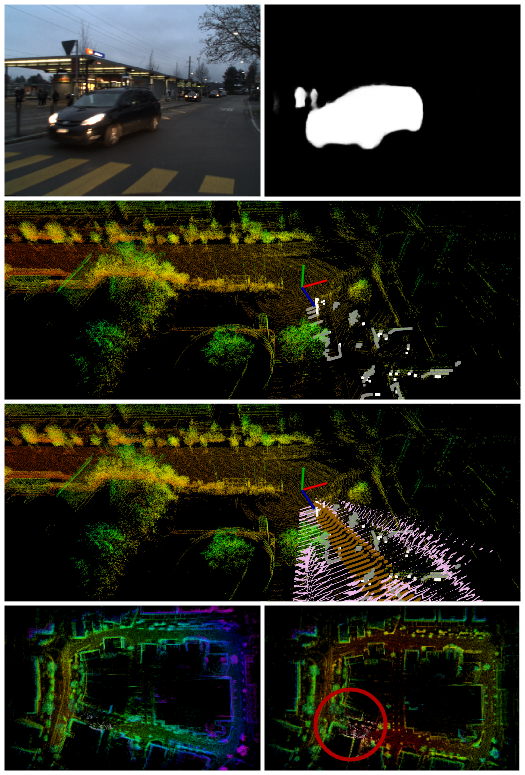}
}
\caption{Pose estimation examples on DSEC zurich\_01\_e: \emph{(top row)} reference image and motion segmentation mask, \emph{(second row)} motion mask projected on LiDAR (static: gray, dynamic: white), \emph{(third row)} projected depth estimate and motion mask from events (static: purple, dynamic: brown), \emph{(bottom left)} accumulated LiDAR from events pose output, \emph{(bottom right)} accumulated LiDAR from LiDAR odometry with highlighted drift at loop-closure. }
\label{fig:slam}
\end{figure}

\subsection{Depth and Pose estimation}
We add the proposed pipeline to an open-loop LiDAR-based pose estimation system, on top of GTSAM~\cite{gtsam} and demonstrate pose estimation of LiDAR-only odometry~\cite{koide2021voxelized} and an auxiliary events-only odometry. The latter serves as a proxy qualitative measure of our pose estimation step. Since DSEC does not provide ground truth poses, we evaluate relative trajectory error with reference to the LiDAR output and list results in Table~\ref{tab:dsec}. We observe that the event odometry deviates on average by 3.73\% and consistently below 5\% from the state-of-the-art LiDAR baseline. Additionally, Fig.~\ref{fig:slam} illustrates qualitative comparisons between the LiDAR-based and the event-based pose estimation pipeline. Please note that for the event-based results the LiDAR point cloud is projected from the individual poses for illustration purposes only and was not used for pose estimation. Qualitative observation of the outputs furthermore showed no critical failures in the event or LiDAR odometry. On our own data, we find that both LiDAR and event odometry have similar performance with respect to RTK-GPS ground truth, see Table~\ref{tab:robo}. Although we do not further evaluate the depth network, we observe that its output is up to scale with the LiDAR data. Finally, it is important to note that while the LiDAR baseline generates more crisp point cloud maps, it is also subject to moderate drift and does not represent an absolute ground truth. As a side note, we remind that our pipeline does not aim for numerical supremacy on the auxiliary tasks (e.g. depth, 6DoF pose, flow estimation), but rather good enough estimates to facilitate the primary motion segmentation step.

\begin{figure}[]
\centerline{
\includegraphics[width=\linewidth]{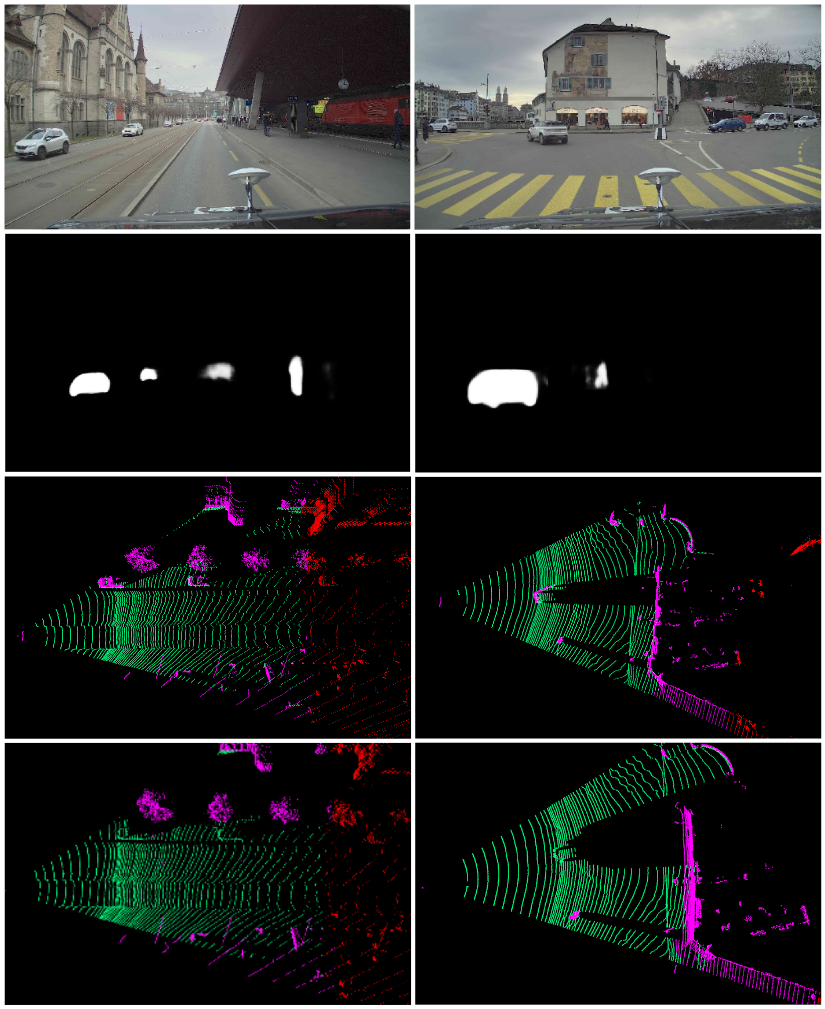}
}
\caption{Drivable Surface segmentation examples on own outdoor data using~\cite{paigwar2020gndnet}. \emph{(top rows)} reference image and motion segmentation masks, \emph{(third row)} Drivable surface segmentation on raw LiDAR point cloud, \emph{(bottom row)} Drivable surface segmentation on motion segmentation masks filtered LiDAR point cloud. The color coding is green: drivable surface, purple: obstacle, red: out of range.}
\label{fig:slam2}
\end{figure}

\subsection{Drivable Surface Detection}
In this demonstration, we filter the LiDAR point clouds on our real driving data (see Section \ref{sec:robo}) by projecting the motion segmentation masks from the event camera on the LiDAR point cloud and cropping it to the overlapping region. Subsequently, we perform Drivable Surface Segmentation on the resulting scans using the segmentation model from~\cite{paigwar2020gndnet} and depict results in Figure~\ref{fig:slam2}. In contrast to the results on un-filtered point clouds, the results indicate that the motion compensation successfully filters dynamic parts which in turn allows for truthful Drivable Surface Detection.

\section{Generation of DSEC-MOTS}

In order to generate segmentation masks for the dynamic objects in DSEC, we first apply a state-of-the-art LiDAR-based dynamic object segmentation algorithm 4DMOS \cite{mersch2022ral}, and then project the LiDAR points to the RGB image plane. Consequently, we run a tracker \cite{pcan} on the RGB images to get their unique ID as well as temporally consistent instance masks. An instance mask is marked as dynamic if the projected LiDAR points hit more than three RGB frames. The instance masks of dynamic objects are then projected to the event plane, based on planar-depth assumption as done in the original DSEC dataset. As a final step, we manually correct the wrong masks. We annotate all the DSEC training sequences (41 sequences, 28179 images) and test sequences (10 sequences, 3956 images) following the official DSEC splits for the left event camera.

\section{Self-collected Outdoor Data} \label{sec:robo}

\begin{figure*}
  \includegraphics[width=\textwidth]{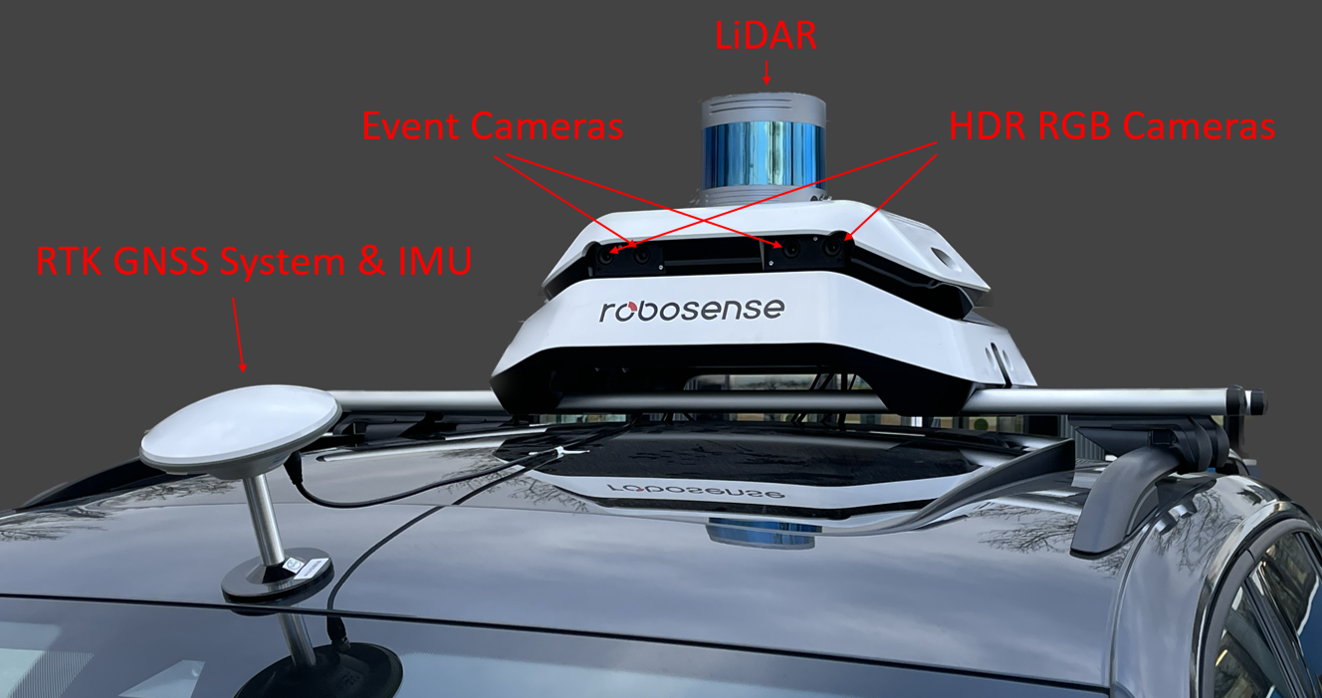}
  \caption{The sensor setup for own data collected in urban environments and used in demonstrating the generalization capabilities of the Motion Segmentation algorithm in the main manuscript.
  }
  \label{fig:collection}
\end{figure*}

We collected data with an augmented version of a commercial driving setup by Robosense, depicted in Fig.~\ref{fig:collection}. The setup consists of a 128 beam RS-RUBY LiDAR, a stereo pair of DVXplorer event cameras at resolution 640 x 480, a stereo-pair of HDR RGB camerasat resolution 3840 x 2160, IMU and an RTK GNSS system. All sensors are hardware synchronized via the GNSS system. We collected 36 sequences in urban environments with a total length of 3 hours and 40 minutes. To maintain anonymity of the submission, we omit further information on the collected data in the initial submission. 

{
    \small
    \bibliographystyle{ieeenat_fullname}
    \bibliography{references}
}